# FanCric: Multi-Agentic Framework for Crafting Fantasy 11 Cricket Teams

Mohit Bhatnagar

Jindal Global Business School, O.P. Jindal Global University, Sonipat, Haryana, India
mohit.bhatnagar@jgu.edu.in

**Abstract.**

Cricket, with its intricate strategies and deep history, increasingly captivates a global audience. The Indian Premier League (IPL), epitomizing Twenty20 cricket, showcases talent in a format that lasts just a few hours as opposed to the longer forms of the game. Renowned for its fusion of technology and fan engagement, the IPL stands as the world's most popular cricket league. This study concentrates on Dream11, India's leading fantasy cricket league for IPL, where participants craft virtual teams based on real player performances to compete internationally.

Building a winning fantasy team requires navigating various complex factors including player form and match conditions. Traditionally, this has been approached through operations research and machine learning. This research introduces the FanCric framework, an advanced multi-agent system leveraging Large Language Models (LLMs) and a robust orchestration framework to enhance fantasy team selection in cricket. FanCric employs both structured and unstructured data to surpass traditional methods by incorporating sophisticated AI technologies.

The analysis involved scrutinizing approximately 12.7 million unique entries from a Dream11 contest, evaluating FanCric's efficacy against the collective wisdom of crowds and a simpler Prompt Engineering approach. Ablation studies further assessed the impact of generating varying numbers of teams. The exploratory findings are promising, indicating that further investigation into FanCric's capabilities is warranted to fully realize its potential in enhancing strategic decision-making using LLM's in fantasy sports and business in general.

## 1 Introduction

Cricket, the world's second most-followed sport, captivates billions of fans with its complex strategies and rich history [1]. The Indian Premier League (IPL), as the pinnacle of Twenty20 cricket in India [2], a format that concludes within a few hours, unlike the longer formats stretching over a whole day (50 overs a side) or multiple days (test cricket), and not only showcases exceptional talent but has also become a central stage for cricket's evolving intersection with technology. As the most popular cricket league globally, the IPL offers a unique window into the dynamic interplay of sport analytics and fan engagement. This study focuses on curating fantasy cricket teams in general with mushrooming multiple franchise leagues across the globe, however our experiments in this research are run within the IPL context, particularly with the Dream11 platform, the most popular fantasy cricket league in India [2]. Participants in Dream11 create virtual teams and are evaluated based on real players' performances, competing in online leagues across the globe, fostering deep online fan engagement.

The challenge of forming a winning fantasy team is multifaceted, influenced by a myriad of factors including player performances, current form, match conditions, home or away game dynamics, and strategic team compositions given several strategic considerations. This complexity makes it an ideal candidate for advanced analytical techniques. Traditionally, methodologies from operations research and machine learning have been employed to optimize team selections, evidenced by a growing literature in the area [1,3, 4, 5, 6, 7]. However, the potential for integrating Large Language Models (LLMs), which have been transformative in other sports domains [8] and especially in context of recent development with respect to autonomous and multi-agent architectures [9, 10, 11, 12] remains largely unexplored.

This research introduces the FanCric framework, a multi-agent system that leverages the capabilities of LLMs using the LangGraph orchestration framework from LangChain [13,14] to navigate the intricacies of IPL fantasy cricket. By harnessing both structured and unstructured data, FanCric aims to outperform traditional fantasy team selection methods. For a specific IPL match, we scraped approximately 12.7 million unique entries from a Dream11 contest, reflecting a vast dataset of individual predictions and strategies. Participants in these contests select a lineup of 11 players, including a captain and vice-captain, whose performances in the actual match determine their fantasy team's success measured by a point-based system.

In this study, we delve into designing and evaluating the FanCric framework, which integrates large language models into the fantasy cricket team selection process. Our research also focuses on comparing the performance of the FanCric system to the collective choices of fantasy sports participants, commonly known as the wisdom of crowds [15, 16]. We also compare results to a simpler prompt engineering approach for team selection to highlight its relative sophistication. Our exploratory findings are positive and highlight the potential of advanced AI applications. These applications, particularly multi-agentic LLMs, are proving valuable in sports analytics and broader business strategy decision-making. This research thus offers a glimpse into the fast-evolving capabilities of AI technology and its expanding impact across domains.

## 2 Literature Review

The field of sports analytics has seen significant evolution over the past decade, with artificial intelligence and machine learning models fast transforming the domain [17]. In cricket, the application of AI, in particular machine learning has facilitated detailed performance analysis and predictive modeling, enhancing decision-making processes in the sport [1]. These advancements and the popularity of the sport have contributed to the rising popularity of Fantasy 11 contests, particularly in leagues built around the Indian Premier League, where fans actively engage in creating and managing their own teams based on real-time data and player performance, with Dream 11 being the most popular platform [2].

Recent studies have focused on various statistical, AI, and optimization techniques to prepare optimal Fantasy 11 teams. For example, a study implemented a team recommendation system using machine learning and skill-based ranking of players, considering multiple factors such as batting, bowling, fielding skills, pitch type, and weather conditions [3]. Optimization techniques such as integer programming have also been applied to maximize the overall performance of Fantasy 11 teams, ensuring a balanced mix of players [4,5]. Furthermore, traditional methods like the Technique for Order of Preference by Similarity to Ideal Solution (TOPSIS) have been employed to evaluate and select the best players based on multiple performance criteria [6]. Notably, a recent study compared different team selection methods using the fantasy league scoring system, highlighting the effectiveness of machine learning and optimization techniques in achieving higher fantasy scores [7].

Large Language Models (LLMs) have shown remarkable potential across a broad range of applications including strategic tasks and as autonomous agents [9]. These models are increasingly used for complex decision-making processes, leveraging their ability to process and analyze vast amounts of data. For instance, LLMs have been employed extensively in board games and video games, demonstrating their strategic and local reasoning capabilities [8]. Studies such as those on k-level reasoning and LLM as a mastermind have explored the application of LLMs in these domains, showcasing their potential to enhance strategic gameplay and decision-making [18, 19]. Prompt engineering is the preferred approach for getting desired results in most applications [20].

Multi-agent systems have off-late shown to improve performance by distributing tasks among multiple agents, each specialized in a particular area [10, 21]. This approach has been effectively applied across various domains such as simulating games [11], software development [12], and financial trading [22]. By breaking down complex tasks into smaller, manageable components, multi-agent systems can optimize processes, enhance efficiency, and achieve better outcomes. These systems leverage collaboration and communication among agents to adapt dynamically to changing environments, making them highly suitable for applications that require complex decision-making, strategic thinking, and planning [23].

The use of LLMs in sports analytics is expanding, with applications in predicting events in various field sports such as football and the NFL. Studies have shown that LLMs can effectively forecast match outcomes and player performance, enhancing the strategic planning capabilities of teams and coaches [24, 25]. Specifically, the Sports Metrics study demonstrates how LLMs blend textual and numerical data to predict player performance and game outcomes, providing a benchmark for assessing numerical reasoning and information fusion capabilities of LLMs in sports analytics [25]. Furthermore, LLMs have also been used to predict fantasy teams in soccer competitions that combines modeling players' fantasy points with subsequent lineup optimization using an ensemble of generative models [26].

Given the above context, there is a need for research to adopt these multi-agentic frameworks for predicting popular fantasy sports leagues, which we do here for cricket. With the availability of open-source frameworks like AutoGen [27] the modeling of the complex and multi-dimensional strategic criterions required for building fantasy teams is easily feasible [27], and in this study we demonstrate the same for curating **Fan**tasy **Cric**ket teams and call our framework in short **FanCric**.

## 3  FanCric Architecture

FanCric architecture leverages an open-source multi-agentic framework LangGraph [13] for its implementation. LangGraph, developed on the LangChain[14] platform, is designed for constructing stateful multi-agent applications using Large Language Models (LLMs). It facilitates the creation of complex agent workflows such as cyclic graph structures, enhancing flow and state management. This framework supports the definition of workflows, flow control, and scalability, making it particularly suited for the FanCric framework, where managing multiple agents and complex decision-making is crucial for optimizing fantasy cricket team selections. Building on this, the construction of a Fantasy 11 cricket team involves a complex decision-making process that must account for various factors to ensure optimal team performance. This includes a deep analysis of player statistics, current form, historical performances, and contextual factors such as match venue, opponent strength, and prevailing weather conditions. Additionally, strategic considerations such as player roles, strengths & weaknesses, team composition, and potential point-scoring opportunities must be meticulously evaluated

The FanCric framework is designed to streamline the process of building Fantasy 11 cricket teams through a coordinated effort of specialized LLM agents, each contributing to different stages of the team selection process. This multi-agent approach is known to enhance the efficiency of the tasks like team assembly in our current study but also ensures that each decision is informed by deep, contextual insights gathered and processed through sophisticated Large Language Model (LLM) interactions. Fig. 1 provides a block diagram for the implementation. The user puts in his query which is routed by the **Supervisor** once the required details from the user is obtained about the fantasy league (e,g. Dream 11), tournament (e.g. IPL), season and the specific match.

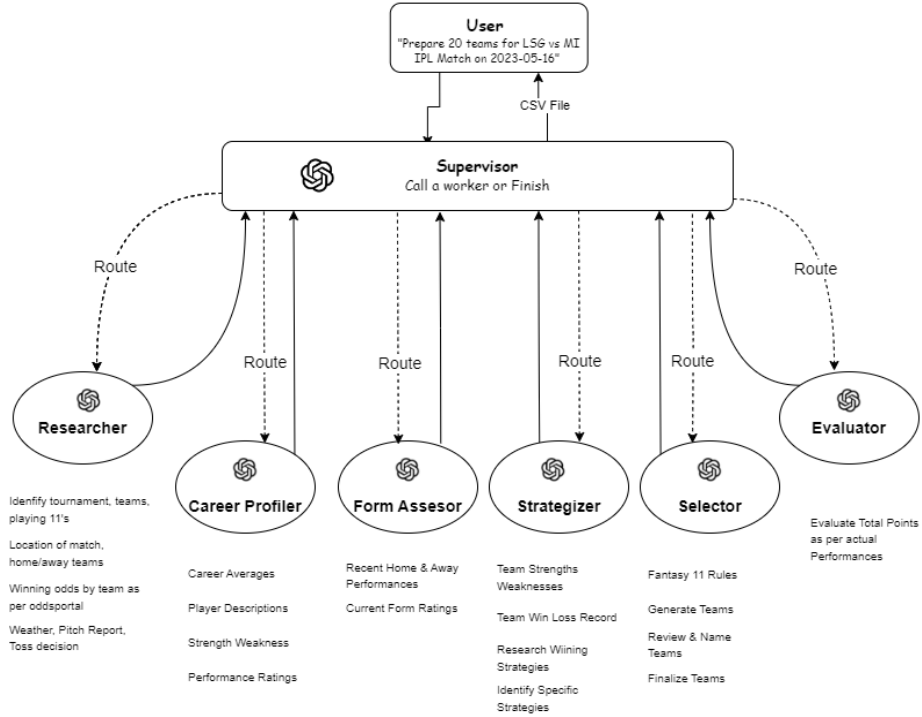

**Fig. 1:** FanCric Multi Agentic Architecture and Descriptions

The **Researcher** agent initiates the team building process by gathering crucial preliminary information about the match. This includes identifying the teams involved, and the roster of players likely to participate in the game. It also assesses the match's venue and determines the home and away teams, gathers data on the weather conditions for the venue using the free API's from https://open-meteo.com/. The weather coordinates are obtained for both the innings separately and includes factors like temperature, wind speed, cloud cover, dew and humidity. Next, we use Tavily, an agentic search tool [28] to collect the odds ratio for winning for the two teams from oddsportal.com. The framework also collects insights on the pitch characteristics using news reports searched again using the agentic search tool. Use of agentic search ensures that the LLM get specific answers with the tool taking care of implementing RAG etc. internally [28]. This foundational data collection sets the stage for informed decision-making in subsequent modules and the information is passed back to the **Supervisor**.

Following the initial research phase, the **Career Profiler** takes over to analyze the historical performance of the shortlisted players in all previous seasons of the tournament. This agent dives into the career statistics of the players, providing a comprehensive review of their past achievements, the descriptions (e.g. right-handed hard hitting opening batsman) and areas of strength and weakness. The LLM capabilities of summarization are used for generating the qualitative descriptions and strengths and weaknesses. It synthesizes this data to rate each player's overall performance relatively and considering the up to date and relevant statistics for a batsman, bowler, all-rounder or wicket-keeper, thus equipping the overall context with a clear understanding of the player capabilities and prior performances.

Subsequently, the **Form Assessor** agent focuses on more recent player performances to gauge their current form. This module tracks the latest match outcomes and performance metrics for each player, considering home/away games, batting first/second thus offering an updated form rating that reflects the player's immediate past performances. This real-time assessment helps in adjusting team selection strategies to current dynamics, making the framework adaptive to the changing conditions (e.g. batting position, injury, form etc.) in the fast paced cricket formats like T20 where reputations are made and broken on a regular basis.

Next, the **Strategizer** agent analyzes the teams' recent performance records, including win-loss ratios in various conditions and outcomes based on batting first or second. It also examines team strengths and weaknesses and uses this analysis to search for and develop effective strategies for the fantasy team composition. A search on the internet and social media sites is also conducted for top tips and recommendations for choosing winning fantasy teams for the specific match or tournament in general. The Strategizer agent based on all the inputs then proposes actionable strategies tailored to the specific match conditions, thus the agent plays a crucial role in steering the team selection towards the most promising configurations.

The **Selector** agent begins by using the Tavily agentic search to scrape and summarize essential fantasy team building rules, such as player quotas from each team and type requirements, ensuring compliance. It then combines player performance data and strategic inputs from other agents to assemble the required number of fantasy teams. This agent also iteratively reviews and refines the team setup, suggesting adjustments based on a holistic evaluation of all gathered data, thereby optimizing the team's potential for success. The agent also generates the logic at this stage for each team and the same is used by the LLM to provide names for each such team generated, thereby providing insights into the selection thesis.

Finally, the Evaluator agent provides a post-match analysis by calculating the fantasy points earned by the different teams. This agent is incorporated by us as a back-test mechanism to evaluate the quality of teams generated for matches already played and for assessing the accuracy and effectiveness of the team selection strategies and contributes to refining the implemented approach. In a production system the Supervisor can run the scoring logic only once the fantasy points are available online at the end of the game.

Together, these agents thus form a robust framework that leverages the power of multiple LLM's and data-driven insights to optimize Fantasy 11 cricket team selections. This system not only simplifies the complex task of team assembly but also enhances decision-making through a scientifically grounded, multi-dimensional analysis approach considering a comprehensive set of qualitative and quantitative factors.

## 4 Experiments

In this section, we embark on a systematic exploration of the FanCric framework through a series of experiments designed to assess its effectiveness and efficiency in constructing Fantasy 11 cricket teams. Our experiments are tailored to evaluate the performance of FanCric against traditional prompting LLM approaches as well as comparing results with the entries from the participants of a fantasy 11 event on Dream11 for a randomly identified IPL game. The different metrics and experimental settings, help us to compare the approaches and also help us to understand the effects of various components in our framework. This experiment also helps in evaluating the FanCric architecture and its predictive and optimization capabilities in real-world settings.

### 4.1 Research Questions

The main research intent of this study is to examine the effectiveness of the FanCric framework in assembling winning Fantasy 11 cricket teams. Our experiments are structured around the following key research questions:

- **RQ1:** How does FanCric's performance compare to teams generated by an LLM, purely using a prompt engineering approach [19]?
- **RQ2:** How does the performance of FanCric's proposed teams measure up against the 'Wisdom of Crowds' team, defined as the mean or median score of all teams participating in a fantasy contest [15, 16]?
- **RQ3:** How effectively do the individual agents within the FanCric framework contribute to the generation of relevant and actionable information for fantasy cricket team selection?
**RQ4:** Through ablation studies, explore the effect of varying the number of generated teams on the performance of the proposed teams, evaluating if the number of teams generated impacts the performance.

Given the nascent state of applying multi-agent systems to fantasy cricket team selection and in general, this study adopts an exploratory approach focused on broad research questions rather than specific hypotheses. This method allows for an open exploration of how various factors influence team performance in Dream 11 fantasy leagues, setting the stage for more targeted hypotheses in future studies.

### 4.2 Experimental Setup

Our detailed experimental analysis was centered on a specific randomly chosen match from the 2023 IPL season, Lucknow Super Giants (LSG) vs Mumbai Indians (MI), played on May 16$^{th}$ as the 63$^{rd}$ of 72 games in the season at the Bharat Ratna Shri Atal Bihari Vajpayee Ekana Cricket stadium in Lucknow. The dataset for comparison of how the team faired comprised of Dream 11 fantasy team selections from over 13.8 million entries, parsed from 14 different PDF files (each having a million entries spread across several thousand PDF pages each). We used the pyspark big data framework and python pdfplumber library for parsing and the code was run on Google Collab Pro+, given the need for high compute resources for the activity. This data is made available on the Dream 11 mobile application for all participants to a contest. For accessing the files an android emulator (https://www.bluestacks.com) was used to transfer the files from the mobile application to a personal computer and subsequently stored in Google drive for processing using the Google Collab environment.

Player performance data up to the 2022 IPL season and the detailed 2023 IPL season performance leading up to the evaluated match was collated from Kaggle [29] and the official IPL website [30]. The prompts were carefully designed to ensure that the LLM (GPT-4o-mini) did not access the specific match data and results for making its choices. Besides prompting as far as feasible identifiers like PlayerID obtained from ESPN Cricinfo[31] was used instead of Player names etc. to ensure unbiased player ratings and selection decisions, thus removing potential bias. Additionally, rigorous data validation processes were implemented to check for inconsistencies or errors in the data set, enhancing the reliability of our analysis and outcomes.

For comparing results, metrics were calculated for individual teams and aggregated when generating multiple teams. At the team-specific level, we analyzed the percentile rankings of total fantasy points scored by each team in the contest. Additionally, we assessed the accuracy of selections for players, captains, and vice-captains in each team relative to the Dream Team (DT), which represents the optimal team configuration. A win count was also considered which was the count of teams for which the entry fees were returned (for the contest being considered this was done for 67% of the entries). At an aggregate level the average total fantasy team score, average hit rate, the win rate and the percentile rank of the best performing team were reported for t generated teams.

### 4.3   Results

In this section, we address the research questions outlined earlier by comparing performances from the Wisdom of Crowds, the prompt engineering approach, and the FanCric approach. Specifically, we evaluate the team selection performance of the prompt engineering and FanCric approaches and compare them to the team chosen by the Wisdom of Crowds. For consistency in our comparisons of the prompt engineering and FanCric approaches, we standardize the evaluations by generating 10 teams for each method. The value of 10 was chosen as it corresponding to the median number of entries allowed on the Dream 11 platform for varying sizes of contests. This approach thus ensures a focused analysis of how each method performs under similar conditions. A more nuanced evaluation is included in the ablation studies section.

**Wisdom of Crowds**

In the examination of the Wisdom of Crowds, applied to choosing Fantasy 11 cricket teams, our data scraping efforts initially collected 13.8 million team entries. Post the removal of duplicates, we were left with 12.7 million unique teams, offering insight into the popular choices and strategies prevalent among participants. Suryakumar Yadav from Mumbai Indians was the most favored captain with over 3.28 million picks.

picks, with Quinton de Kock from Lucknow Super Giants as the top Vice Captain with 1.66 million picks. The other 9 players were similarly selected in the Wisdom of Crowds teams and represented the most chosen picks in the various teams, reflecting the collective preferences of the fantasy league participants. Analyzing the total Fantasy League points, the teams amassed an average of 501.54 points with a standard deviation of 89.02, spanning from a minimum of 0 to a maximum of 811.5 points, with the median at 512 points.

The distribution of total scores, as illustrated in the attached histogram (see Fig.2), was notably left-skewed, indicating a concentration of teams achieving higher scores, with fewer teams gathering lower scores. This distribution failed the normality test as evidenced by a Kolmogorov-Smirnov statistic of 0.14 and a p-value of 0.0, confirming the non-normal distribution of the data. The 'Dream Team', representing the theoretical maximum score achievable, registered 832.5 points, which was 21 points higher than the maximum score achieved by any participant, despite the extensive number of entries.

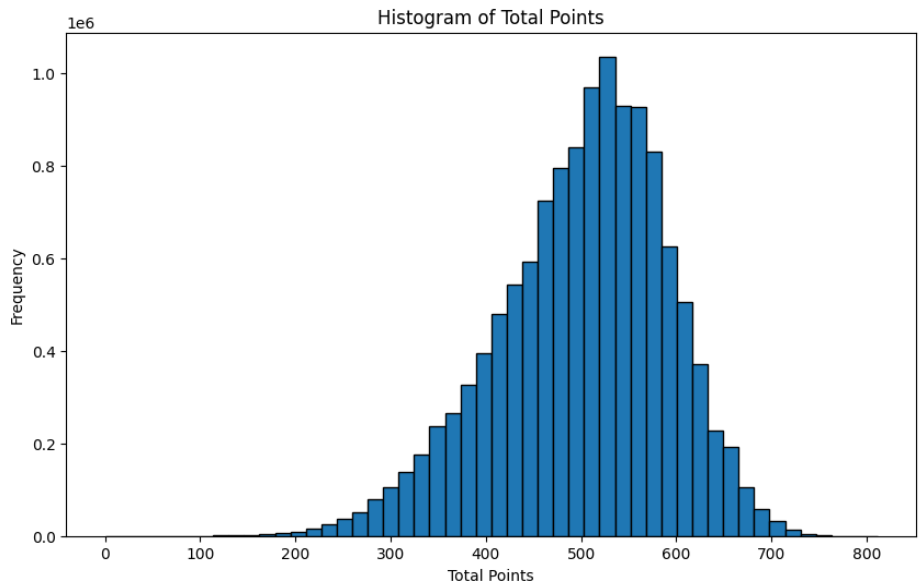

**Fig. 2**: Distribution of Total Points earned by the Wisdom of Crowds team

**Prompt Engineering Approach**

Table 1 elucidates the outcomes of the prompt engineering approach applied to generate Fantasy 11 cricket teams through ChatGPT API, showcasing the performance of teams headed by varied captains along with their respective scores and rank percentiles. The data ranges from a low score of 455.5 points at the 27.9 percentile to a high score of 634 points at the 95.1 percentile. A few shot prompt approach, which included two example responses was used to generate the teams using the Chat Completions API from OpenAI for the gpt-4o-mini model. The default parameter settings (temperature = 1) were used with the specific prompt being *"You are a cricket analyst. For an IPL match between {teamA} and {teamB} at {city}, please generate 10 Fantasy 11 teams based on the following 22 players {playerTeamA}, {playerTeamB} which can participate on Dream 11 as per its rules and also have a high chance of winning the Dream 11 contest. You have to choose a Captain, Vice-Captain and 9 players. Example ... snip ..."*.

The average score for these teams stood at 512.9, placing them at the 50.4 percentile, with a notable 70% win rate, indicating a modest yet effective outcome from this computational approach. This average score slightly edges out the average of 501.54 points derived from the Wisdom of Crowds approach, where a vast dataset of 12.7 million unique teams was analyzed. Despite the narrow margin, the results underscore the potential of guided, strategic prompt engineering in generating competitive fantasy teams, particularly in contrast to crowd-sourced strategies which Dream 11 participants adopt today (building teams derived from experts, called Guru teams).

**Table. 1:** ChatGPT generated teams by Prompt Engineering

| S.No. | Team | Total Points (Rank Percentile) | Hit Rate with DT (C, VC, P) | | | Win (Y/N) |
|---|---|---|---|---|---|---|
| 1 | Quinton de Kock-C... | 507(47.8) | 0 | 0 | 7 | Y |
| 2 | Suryakumar Yadav-C... | 455.5(27.9) | 0 | 1 | 5 | N |
| 3 | Rohit Sharma-C... | 495(42.8) | 0 | 0 | 6 | Y |
| 4 | Marcus Stoinis-C... | 602(88.4) | 1 | 0 | 7 | Y |
| 5 | Deepak Hooda-C... | 459.5(29.2) | 0 | 0 | 6 | N |
| 6 | Ishan Kishan-C... | 524(56.3) | 0 | 0 | 6 | Y |
| 7 | Tim David-C... | 634(95.1) | 0 | 0 | 8 | Y |
| 8 | Quinton de Kock-C... | 487.5(39.9) | 0 | 0 | 7 | Y |
| 9 | Rohit Sharma-C... | 458.5(28.9) | 0 | 0 | 5 | N |
| 10 | Nicholas Pooran-C... | 506(47.4) | 0 | 0 | 7 | Y |
| | Average, Win Rate | 512.9(50.4) | 0.1 | 0.1 | 6.4 | 70% |

**FanCric Approach**

The results for the various orchestrated interactions in summarized in Fig.3. and the interactions highlight the decision-making process the agents followed for building the 10 optimized fantasy 11 cricket teams. The Researcher agent initially provided the game setup and details like the toss winning team, decision to bat/bowl first and the playing 11s. Since in a Dream 11 contest a team can be edited and finalized till the match starts, this was a simulation of the real-life scenario as well when the toss decision etc. is available to the contest participants. The agentic search provided the pitch conditions at the venue and indicated that the pitch typically leads to low-scoring games with equal contest between batting and bowling, thereby setting the stage for strategic team formulations.

Following the initial assessment, the Career Profiler considered the player past performances to relatively rate them and this also considered the descriptions of the player. In Fig.3, we highlighted how Marcus Stoinis, who was a significant contributor in the game as player of the match and also the player who won most points, was evaluated by the FanCric framework. The evaluation of his strengths and weaknesses highlighted his role as a high-impact all-rounder. His overall career rating of 7 was complemented by recent performance scores of 8 out of 10 from the form assessor. This pinpointed evaluation highlighted his potential crucial role in the game, an insight that was vital for constructing competitive fantasy teams where he was chosen as a Captain and Vice-Captain in 2 of the 10 teams generated by FanCric.

The Strategizer analyzed recent win-loss records, noting that the Lucknow Super Giants (LSG) had inconsistent results, while the Mumbai Indians (MI) showed a stronger winning record. However, with the odds being nearly equal, it emphasized the importance of selecting players from both teams to maintain balance. It then conducted a strength & weakness analysis of the two teams revealing LSG's robust batting capabilities contrasted by inconsistencies, particularly in the middle order and under pressure situations. For MI, a formidable batting lineup and seasoned bowling pose significant advantages, however it was highlighted that they struggle with consistency in the middle order and fielding.

The Strategizer also considered general strategies and tips that are popular on social media for crafting competitive IPL Dream11 teams. It identified key trends like prioritizing all-rounders and top-order batsmen, which are vital given the short form of the game where only a few get opportunities. Based on this data and specific match conditions, the Strategizer outlined strategic recommendations. For example, it suggested emphasizing top-order batsmen from LSG due to their stronger batting lineup, home conditions and the historically challenging pitch conditions, which are likely to favor bowlers, particularly spinners in the second innings.

The Selector agent played a pivotal role in generating and refining the Dream11 fantasy team compositions by integrating feedback. After initial team proposals were generated based on the match conditions, player evaluations, strategies and fantasy rules the Selector proposed the 10 teams that were evaluated by the review and feedback mechanism where we used GPT4o (a more advanced model that could review and provide a feedback) instead of GPT4o-mini which was used for simulating all other agents. This iterative process ensured that each team not only complied with Dream11's regulations but also maximized competitive advantage by optimizing player selection. Feedback focused on enhancing team balance, leveraging player form, and adhering to strategic insights such as exploiting specific player known match winners as the Captain and ViceCaptain and adjusting to other game conditions. The loop of feedback and adjustment enabled better team selection was verified as the Finalized Teams had better choices for Captain and Vice Captains than the initial ones proposed. This methodical refinement thus helped in crafting teams that were rule-compliant and also strategically poised to score high in the fantasy league. The agent was also prompted to generate a rational for the chosen team and short team name based on the same in alignment to the Chain of Thought (COT) reasoning approach in prompting [33].

Finally, each of the teams generated were evaluated based on the dream 11 points earned for the game by different players playing. Further evaluation of the FanCric approach, detailed in Table 2, demonstrates the effectiveness of the multi-agentic strategy in producing an average score (528.55) which stood at 58.6 percentiles when compared with the Wisdom of Crowds approach and was also better than the prompt engineering approach. The results thus point towards the benefits of the FanCric approach. The "Strategic Strikers" team, which included Marcus Stoinis as Captain and also had 6 of the 11 dream team players, performed the best, scoring the 96th percentile among all 10 entries. This evaluation underscores the efficacy of the multi-agent systems in leveraging both data-driven insights and strategic adaptability to include qualitative factors and deliberations to reach their decisions.

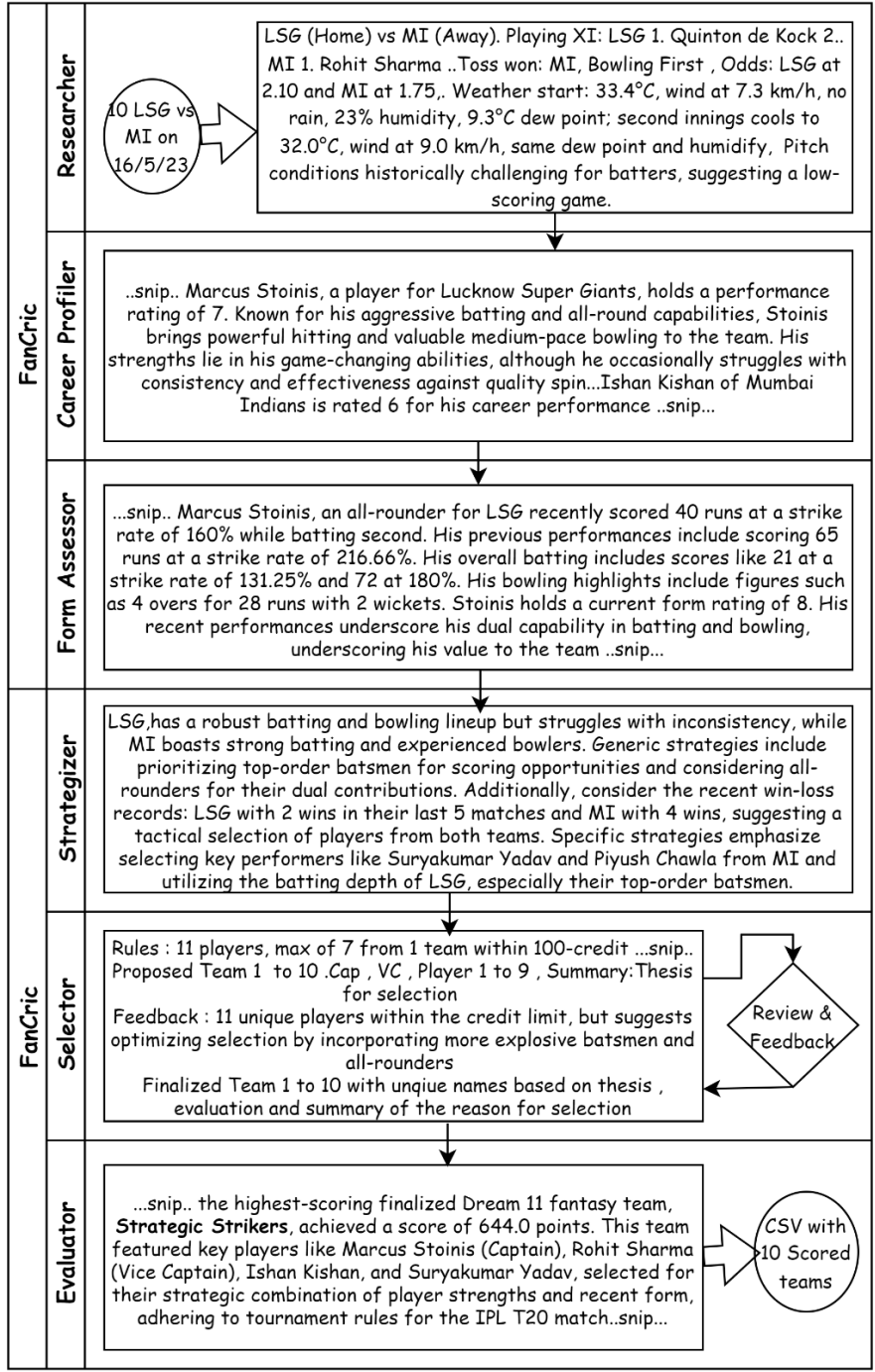

**Fig.3.** FanCric Agent Interactions

Table. 2: ChatGPT generated teams by FanCric

| S.No. | Team | Total Points (Rank Percentile) | Hit Rate with Dream Team C, VC, P | | | Win (Y/N) |
|---|---|---|---|---|---|---|
| | Rohit Sharma -C... | 523.5(56.1) | 0 | 0 | 6 | Y |
| | Suryakumar Yadav -C... | 558.5(72.4) | 0 | 1 | 7 | Y |
| | Nicholas Pooran-C... | 543(65.3) | 0 | 0 | 7 | Y |
| | Marcus Stoinis-C... | 644(96.3) | 1 | 0 | 6 | Y |
| | Ishan Kishan-C... | 497.5(43.8) | 0 | 0 | 6 | Y |
| | Ayush Badoni -C... | 405(14.7) | 0 | 0 | 5 | N |
| | Chris Jordan-C... | 535(61.7) | 0 | 0 | 7 | Y |
| | Ravi Bishnoi -C... | 455(27.8) | 0 | 0 | 5 | N |
| | Jason Behrendorff -C... | 572(78.6) | 0 | 0 | 6 | Y |
| | Yash Thakur -C... | 552(69.4) | 0 | 0 | 6 | Y |
| | Average, Win Rate | 528.55(58.6) | 0.1 | 0.1 | 6.1 | 80% |

### 4.4  Ablation Studies

The ablation study conducted examined the effect of varying the number of fantasy cricket teams (n) generated on the performance in Dream11 contests, where n was chosen as 1, 5, 10, 15, and 20. This selection range is reflective of Dream11's contest rules, where the number of permissible team entries are limited to a maximum of 20 for larger contests and varies between 1 and 20 for smaller ones. The results which include the average points scored by the teams, percentile rank averages, hit rate in selecting captains or vice-captains and players from the Dream Team, the win percentage indicating teams that ranked within the top 66.7 percentile and the top ranked team in the set are presented in Table 3. The outcomes from a prompt engineering approach are included in brackets as a benchmark for comparison.

In the analysis, it is observed that the FanCric approach generally yielded higher average points compared to the prompt engineering approach, except when only one team was generated. This indicates the inherent unpredictability's in cricket and fantasy team selection - the advanced analytics employed by FanCric begin to show it's effectiveness only as the number of generated teams increases. This trend is evident from the higher average points and percentile ranks recorded for larger sets of generated teams. Although the prompt engineering approach selected Dream Team players on an average more, however this also indicates that the FanCric approach led to player selections that scored higher overall, suggesting a more effective selection of players.

The win rates between the two approaches were comparable, but notably improved in the FanCric model as the number of teams increased. A standout result in the study was a FanCric team achieving a 99.93 percentile rank with a total score of 722 when 20 teams were generated, underscoring the potential of a multi-agentic approach to enhance outcome predictions as more options are generated. The FanCric approach consistently outperformed the average (501.5) and median scores (512) of the Wisdom of Crowds, except in the n=5 case. This enhanced performance in team selection by FanCric as n increases, thus suggests that generating a larger number of teams does not degrade performance and on the contrary can increase the likelihood of selecting optimal captains, vice-captains, and Dream Team members, thereby maximizing fantasy league success where only a very few entries are rewarded handsomely.

**Table. 3:** Ablation studies for FanCric (Prompt Engineering)

| n | Points Avg, | Rank Avg. | C/VC in DT | Players in DT | Win % | Highest Rank |
|---|---|---|---|---|---|---|
| 1 | 522.5 (535.5) | 55.6 (61.9) | 0 (0) | 7 (7) | 100% (100%) | 55.6 (61.9) |
| 5 | 507.7 (501.7) | 53.0 (55.1) | 0 (0.2) | 6.2 (6.6) | 60% (80%) | 86.3 (86.1) |
| 10 | 528.6 (512.9) | 58.6 (50.4) | 0.1 (0.1) | 6.1 (6.4) | 80% (70%) | 96.3 (95.1) |
| 15 | 517.9 (498.4) | 54.1 (49.1) | 0.1 (0.1) | 5.9 (6.6) | 66.7 (66.7) | 99.1 (99.3) |
| 20 | 528.5 (481.6) | 56.9 (42.9) | 0.15 (0.13) | 5.7 (6.3) | 75% (50%) | 99.9 (95.9) |

## 5 Discussion

The analysis of the FanCric framework provided evidence regarding its efficiency in assembling fantasy cricket teams, evaluated against established benchmark of the wisdom of crowds [15.16]. For RQ1, FanCric outperformed the standard LLM (ChatGPT) prompt engineering approach in most studies, particularly as the number of generated teams increased. This suggests that the multi-agentic, nuanced approach of FanCric is superior for complex decision-making tasks such as fantasy cricket team selection, where variability and unpredictability are driven by identifiable factors. However, the performance benchmarks could be further improved with further fine tuning the FanCric approach to improve its performance. In addressing RQ2, our findings showed that FanCric's teams consistently exceeded the performance metrics established by the Wisdom of Crowds, indicating its capability to leverage and synthesize data effectively and indicating a potential application in fantasy cricket leagues.

Regarding RQ3, the cohesive operation of individual agents within the FanCric framework, the Researcher, Career Profiler, Form Assessor, Strategizer and Selector proved crucial. As we saw in the qualitative assessment each agent contributed uniquely to refining the selection process, emphasizing the value of integrating diverse analytical and qualitative perspectives in sports analytics. It also indicates the value of human like deliberations possible between LLM's with multi-agentic frameworks for decision making.

Our ablation studies for investigating RQ4, highlighted that increasing the number of teams generated directly correlates with no discernible decrease in performance however can improve chances of winning. This relationship suggests that higher quantities of generated teams enable a broader exploration of potential team configurations, which in turn increases the likelihood of approaching an optimal team setup. Particularly, a FanCric team with 20 entries achieved the 99.93 percentile, illustrating the potential of using a more extensive base of generated teams to refine and perfect fantasy team selections.

## 6      Conclusions and Further Research

This exploratory study with the FanCric framework has shown promising results in the context of fantasy cricket team selection. While FanCric has demonstrated potential advantages over traditional methods and the Wisdom of Crowds approach, the findings which are based on retrospective analysis necessitate further empirical studies to more robustly confirm its superiority for live games in real-life settings. The current study highlights FanCric's initial efficacy, but additional simulation studies are crucial to establish its effectiveness conclusively.

Future research should focus on refining the FanCric architecture, incorporating advanced AI/ML algorithms and operations research techniques into the framework that could enhance decision-making processes further. This might involve adapting the system to integrate other insights like social media opinions, expert opinions etc. which could more dynamically adjust team selections based on pertinent game insights. Moreover, expanding the scope of ablation studies are needed to identify which components of the LLM or specific configurations yield the best results and also understand how varying these to use different LLM's or newer models with improved reasoning capabilities change outcomes. This framework could also be useful to adopt for other sports as well.

Further exploration of the multi-agentic approach as adopted in this study not only promises significant improvements in strategic decision-making within sports analytics but also holds the potential to revolutionize broader applications such as stock portfolio management, marketing strategy optimization, and resource allocation across various business contexts. As highlighted by a recent McKinsey study, integration of agent-based systems in generative AI could fundamentally transform business approaches and decision-making processes [35]. This adaptation could lead to more dynamic and responsive strategies across industries, effectively leveraging advanced AI capabilities to address complex and varied challenges.